# Deep Learning Inversion of Electrical Resistivity Data

Bin Liu, Qian Guo, Shucai Li, Benchao Liu, Yuxiao Ren, Yonghao Pang, Xu Guo, Lanbo Liu, and Peng Jiang, *Member, IEEE*

*Abstract*—The inverse problem of electrical resistivity surveys (ERSs) is difficult because of its nonlinear and ill-posed nature. For this task, traditional linear inversion methods still face challenges such as suboptimal approximation and initial model selection. Inspired by the remarkable nonlinear mapping ability of deep learning approaches, in this article, we propose to build the mapping from apparent resistivity data (input) to resistivity model (output) directly by convolutional neural networks (CNNs). However, the vertically varying characteristic of patterns in the apparent resistivity data may cause ambiguity when using CNNs with the weight sharing and effective receptive field properties. To address the potential issue, we supply an additional tier feature map to CNNs to help those aware of the relationship between input and output. Based on the prevalent U-Net architecture, we design our network (ERSInvNet) that can be trained end-to-end and can reach a very fast inference speed during testing. We further introduce a depth weighting function and a smooth constraint into loss function to improve inversion accuracy for the deep region and suppress false anomalies. Six groups of experiments are considered to demonstrate the feasibility and efficiency of the proposed methods. According to the comprehensive qualitative analysis and quantitative comparison, ERSInvNet with tier feature map, smooth constraints, and depth weighting function together achieve the best performance.

Manuscript received April 9, 2019; revised September 5, 2019 and December 31, 2019; accepted January 18, 2020. This work was supported in part by the grants of National Natural Science Foundation of China under Grant 51739007 and Grant 61702301, in part by the United Fund of National Natural Science Foundation of China under Grant U1806226, in part by the Development Program of China under Grant 2016YFC0401805, in part by the Key Research and Development Plan of Shandong Province under Grant Z135050009107, in part by the Interdiscipline Development Program of Shandong University under Grant 2017JC002, and in part by the Fundamental Research Funds of Shandong University. *(Corresponding author: Peng Jiang.)*

Bin Liu is with the School of Qilu Transportation, Shandong University, Jinan 250100, China, also with the Geotechnical and Structural Engineering Research Center, Shandong University, Jinan 250100, China, and also with the Data Science Institute, Shandong University, Jinan 250100, China (e-mail: liubin0635@163.com).

Qian Guo and Xu Guo are with the Geotechnical and Structural Engineering Research Center, Shandong University, Jinan 250100, China (e-mail: hhu_gq@163.com; guoxu1014@163.com).

Shucai Li is with the School of Qilu Transportation, Shandong University, Jinan 250100, China, and also with the Geotechnical and Structural Engineering Research Center, Shandong University, Jinan 250100, China (e-mail: lishucai@sdu.edu.cn).

Benchao Liu, Yuxiao Ren, Yonghao Pang, and Peng Jiang are with the School of Qilu Transportation, Shandong University, Jinan 250100, China (e-mail: sduliubenchao@163.com; ryxchina@gmail.com; yonghao_pang@163.com; sdujump@gmail.com).

Lanbo Liu is with the Department of Civil and Environmental Engineering, University of Connecticut, Mansfield, CT 06269 USA (e-mail: lanbo.liu@uconn.edu).

## I. Introduction

ELECTRICAL resistivity survey (ERS), as one of the most well-known and commonly applied investigation technologies [1], has been widely used in environmental investigations [2]–[4], engineering prospects [5]–[7], hydrological surveys [8]–[11], and mining applications [12]–[14]. Geological interpretations using observed data are usually far from revealing the complex characteristics of subsurface properties. Thus, to satisfy geological interpretation purposes, geophysical inversion methods are continually focused on reconstructing more accurate and detailed subsurface properties.

After years of development, nonlinear optimization methods have been widely adopted in ERS inverse problems, such as genetic algorithm [15], [16], simulated annealing algorithm [17], and particle swarm optimization (PSO) [18]. Particularly, instead of optimizing resistivity model, artificial neural networks (ANNs) build the mapping between apparent resistivity data and resistivity model directly by updating parameters of networks. With ANNs, some significant results in ERS inversion have been gained in both synthetic and field tests [19]–[22]. Usually, ANNs are optimized by gradients derived from loss functions. Due to the gradient vanish/explosion problems, it is not easy to train a deep and large ANNs with strong modeling capacity at the beginning. Thus, some limitations such as slow convergence, low accuracy, and overfitting phenomenon in training remain [23].

Recently, many approaches are proposed to help backpropagate gradients efficiently through the whole networks, such as residual units proposed by He *et al.* [24], activation functions proposed by Maas *et al.* [25] and normalization layer proposed by Ioffe and Szegedy [26]. With them, ANNs with deep layers and tremendous parameters could be optimized. Accordingly, people tend to refer to deep ANNs as deep neural networks (DNNs). Methods based on DNNs are usually called deep learning which has shown superior performance in many problems that require the perception and decision abilities of machines [27]. Initially succeeding in computer vision for the tasks of image perception [24], [28]–[30], now DNNs have been prevalent in many fields such as computer graphics and natural language processing (NLP). Besides, at present, DNNs have developed many variants with different computational logic, such as Multilayer Perceptrons (MLPs, also known as fully connected networks), convolutional neural networks (CNNs), and recurrent neural networks (RNNs), which further extends the application scenarios.

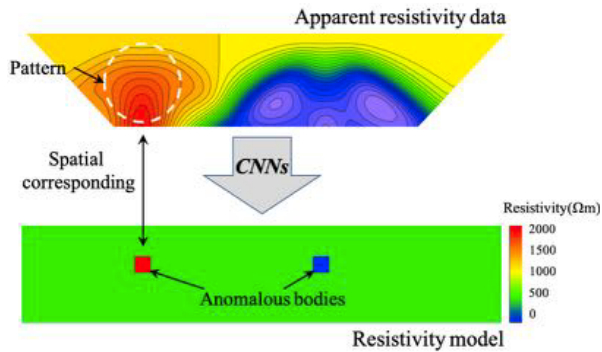

Fig. 1. Task definition. Observed data caused by different resistivity anomalies in the model show different patterns in apparent resistivity pseudosection, and the patterns show certain spatial correspondence to resistivity anomalies.

Many attempts have already verified that DNNs could approximate very complex nonlinear mapping functions, especially for ill-posed inverse problems, such as image superresolution [31], medical imaging [32], and 3-D model reconstruction [33]. The successes have also inspired many approaches for geophysical problems, particularly in the field of seismic inversion. Araya-Polo *et al.* [34] used a velocity-related feature cube transferred from raw seismic data to generate the velocity model by CNNs, whereas Wu *et al.* [35] treated seismic inversion as image mapping and build the mapping from seismic profiles to velocity model directly. Furthermore, Li *et al.* [36] figured out the weak spatial correspondence and the uncertain reflection–reception relationship problems between seismic data and velocity model and propose to generate spatially aligned features by MLPs at first. The latter two works could build the mapping from raw seismic data to velocity model directly without data preprocessing. All of these works demonstrate promising performance in result accuracy and computation speed, which bring new perspectives for ERS inversion. Besides, by utilizing big data, DNN-based inversion methods are more robust and have the potential of the practical application.

In this article, deep learning inversion of ERS is to learn the mapping from input (apparent resistivity data) to output (resistivity model) directly by CNNs, as illustrated in Fig. 1. Typically, the existence of resistivity anomalies in the model will cause responses in apparent resistivity data. Meanwhile, as can be seen from Fig. 1, the responses of observed data caused by different resistivity anomalies also show different patterns, and the patterns demonstrate certain spatial correspondence to resistivity anomalies in the model. As for the patterns in resistivity data, they only cover a part of region, which means they are local existing. In conclusion, the input and output for resistivity inversion have two characteristics: 1) have the spatial corresponding and 2) have the local existing patterns. So, the input and output can be considered as the natural images, and the task can be treated as the common mapping between images. In this case, CNNs are preferred among a variety of DNN variants because they are much powerful in extracting local patterns and more efficiency regarding the number of parameters. Also, the performance of CNNs has been widely concerned in the research of remote sensing scene [37]–[39].

Apart from the similarities with the natural images, apparent resistivity data have their own characteristics. As demonstrated in Fig. 2, when the same anomalous body is located at different vertical positions, the apparent resistivity data show patterns with large differences, namely patterns have the vertically varying characteristic and are not spatially invariant. Meanwhile, the CNNs are with the local and weight-sharing convolutional kernels, which result in a certain receptive field and effective area for CNNs. In this case, when applying CNNs to the apparent resistivity data, there may be situations where CNNs are requested to give different outputs from similar patterns within the effective area. This situation poses a great challenge for CNNs and will make outputs of CNNs ambiguity. The details will be discussed in Section III. This may be the main difficulty for applying prevalently used CNNs to inverse electrical resistivity data.

We adopt the prevalent CNN-based U-Net architecture [40] to design our networks (ERSInvNet). To capture the potential global resistivity distribution change caused by resistivity anomalies, we set ERSInvNet with 30 layers to have enough receptive fields. Then to reduce the potential ambiguity caused by the vertically varying characteristic of apparent resistivity data, we supply ERSInvNet with vertical position information by concatenating an additional tier feature map to the input data. To address the common problem that deep anomalies are difficult to inverse well, we introduce depth weight function in loss function to let network pay more attention to the deep region of resistivity models. Besides, we apply smooth constraints in the loss function to suppress potential false anomalies. In experiments, synthetic examples in our proposed ERSInv data set are used to verify the feasibility and efficiency, and ERSInvNet is trained end-to-end without any data processing. Through comprehensive qualitative analysis and quantitative comparison, the proposed ERSInvNet consistently achieves promising performance regarding the fast inference speed and high inversion accuracy.

## II. BACKGROUNDS

The basic measurements of ERS are made by injecting current into the ground through two current electrodes and measuring the potential difference between other pairs of electrodes. In a typical scenario of ERS, the potential difference data are acquired at the Earth's surface as observed data $d$. The inverse problem involves inferring a set of parameters in model $m$ from a set of data $d$, and it usually relies on the minimization of the objective function. Since we achieve resistivity inversion by building the mapping from the observed data to the resistivity model directly through CNNs, the mechanism of CNNs and related concepts are first displayed in this section.

CNNs are designed for processing natural images which have the local and spatially invariant patterns. For example, in Fig. 3, the pentagram is a graph with a pattern of five-pointed star. This pattern is composed of local structures, and no matter where the pattern is, it corresponds to a pentagram. To make the best use of these characteristics, CNNs with local spatial and weight-sharing convolutional kernels are proposed.

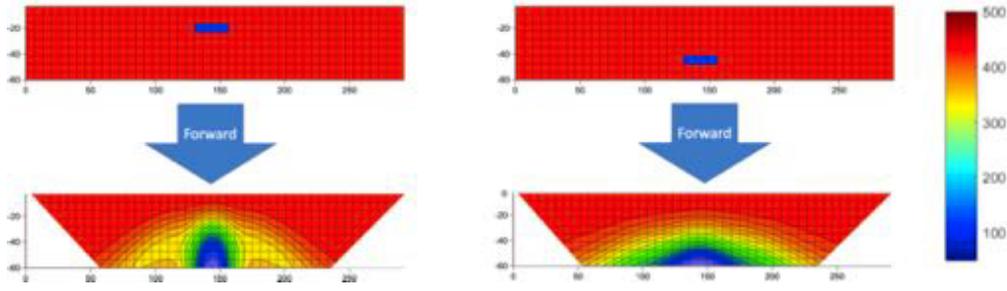

Fig. 2. Two apparent resistivity data and resistivity model pairs. The two resistivity models have the same anomalous bodies but in the different vertical positions. The corresponding patterns of the responses in observed data demonstrate large differences.

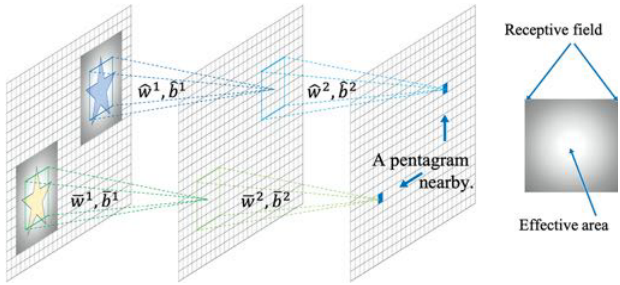

Fig. 3. Illustration of convolutional operation and receptive field concept. The kernel and bias in each layer are weight sharing. The gray area in the input indicates the receptive field caused by two convolutional operations with the $5 \times 5$ kernel for elements in the output. The center parts of receptive field (*effective area* [41]) usually have more influence than the surrounding parts and are indicated with light gray color.

A three-layer CNNs with only one channel in each layer is shown in Fig. 3; we use $w^l$ and $b^l$ to denote convolutional kernel and bias in the $l$th layer, and use $\hat{\cdot}$ and $\bar{\cdot}$ to represent the convolution operations applied on the two different positions. According to the definition of CNNs, $\hat{w}^l = \bar{w}^l$ and $\hat{b}^l = \bar{b}^l$ are referred to as weight sharing properties of CNNs. So that a convolutional layer can be expressed as

$$y^l = f(w^l * y^{l-1} + b^l) \quad (1)$$

where $*$ represents convolutional operation, and $f$ denotes nonlinear activation function. $w^l$ and $b^l$ are, respectively, the convolutional kernel and bias for layer $l$. $y^{l-1}$ and $y^l$ are the input and output of layer $l$. Usually, each convolutional layer could have many kernels ($w^{l,k}, k = 1, \ldots, K$) and will result in output feature of $K$ channels. Thus, when $y^{l-1}$ has $C$ channels, the value in the $k$th channel and position $(i, j)$ of the output in layer $l$ is

$$y^l_{k,i,j} = f\left(\sum_{c=1}^{C}\sum_{m=1}^{M}\sum_{n=1}^{N} w^{l,k}_{c,m,n} y^{l-1}_{c,i-\lceil M/2\rceil+m, j-\lceil N/2\rceil+n} + b^{l,k}\right) \quad (2)$$

where $w^{l,k}$ is the designed $k$th kernel of size $[C, M, N]$ (indicates channel, height, and width of the kernel, and the kernel should have the same channel as $y^{l-1}$) to be applied to $y^{l-1}$. The output of each convolutional layer is the nonlinear weighted of the output of its former layer and is obtained easily. Note that 2 is the most basic 2-D convolutional operation without stride and dilation choice.

Modern CNNs are with many convolutional layers, and each layer includes convolution, nonlinear activation, and some optional operations, such as batchnorm and pooling. Usually, the more the layers CNNs have, the more the nonlinear activation functions are used and more parameters the CNNs would have, so that much stronger the capacity of CNNs is. One rule to decide the number of CNN layers is making them have enough size of the receptive field and meanwhile have enough nonlinear expressiveness.

As for the receptive field, it is the region of neighborhood pixels or grids in the input image the CNNs consumed to give output. As illustrated in Fig. 3 to classify the pentagram of size $7 \times 7$, we should guarantee that each final output element could have the receptive field of size $7 \times 7$ at least and thus be aware of the existence of the whole pentagram nearby. Consequently, we concatenate two convolutional layers with $5 \times 5$ kernels (stride 1 and dilate 1) to satisfy the requirement. Besides, according to Luo *et al.* [41], the center parts of receptive field usually have more influence than the surrounding parts, which is commonly referred to as an *effective area*, which is also marked in Fig. 3.

Certainly, there are many choices to satisfy the requirement of the receptive field, and the size of the receptive field will be affected by many operations such as pooling and upsampling. As for nonlinear expressiveness of CNNs, it is hard to know whether it is enough for the task as the mechanism of CNNs has not been studied comprehensively, so people usually use excessive layers to guarantee the nonlinearity.

## III. METHODOLOGY

### A. Approach

In this article, we intend to learn the mapping function $\mathcal{F}$ from apparent resistivity data $d$ to resistivity model $m$ directly by DNNs that

$$m = \mathcal{F}(d). \quad (3)$$

As stated in Section I, there are several potential problems when using CNNs on apparent resistivity data. First, as illustrated in Fig. 4(a) and proved by Luo *et al.* [41], even using deep CNNs with receptive field cover the whole apparent resistivity data, the patterns within center parts of receptive field usually have much more influence on the corresponding output model value. Thus, the center parts of the receptive field are usually referred to as *effective area*. Second, due to the definition of apparent resistivity, the patterns within the effective area in apparent resistivity data have vertically varying characteristic. There may be situations as shown in the top two figures in Fig. 4(b). Similar patterns within the effective area appear at different tier positions but correspond to different anomalous bodies and model values. Consequently, during training, CNNs may get ambiguous when requested to give different model values from similar input patterns within the effective area. As a comparison, for natural images, as the bottom figure in Fig. 4(b) shows, the patterns of cats correspond to the same semantic meaning in the output no matter where they are located, which means natural images do not have position varying characteristic.

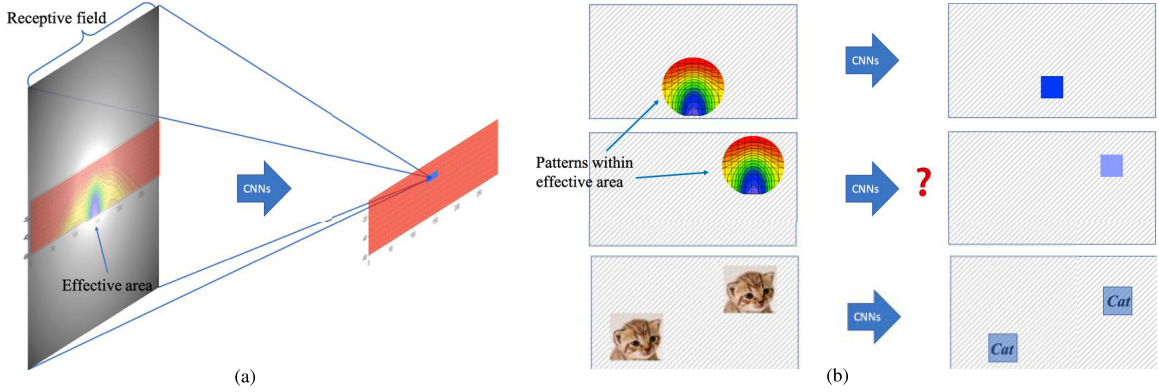

Fig. 4. Illustration of the problem when using CNNs in the raw apparent resistivity data. (a) With deep CNNs, the element in the output will have receptive field cover the whole input data, while the patterns within center parts of receptive field (*effective area* [41]) have much more influence. (b) Different from natural images as shown in the bottom figure, similar patterns within effective area (from the top two figures) may appear at different locations in apparent resistivity data, but with different corresponding model value.

Characteristics of natural images and weight-sharing property of CNNs make CNNs powerful and efficient in dealing with natural images. To make the best use of CNNs in ERS and reduce potential ambiguity, we should supplement input data with more distinguishable information which related to input data, thus reducing the potential ambiguity of CNNs when giving output. In surface ERS, when the distance between two injecting electrodes enlarged, the apparent resistivity data with deeper tier positions in the vertical direction could be calculated to achieve the electrical sounding purpose. The data with deeper tier position have the stronger correlation with the deeper anomalous body, which means that the tier position information is helpful for CNNs to distinguish the data patterns caused by anomalous bodies with different depths. Therefore, adding the tier position information of the data to the input would benefit CNNs for building the mapping. As shown in Fig. 5, CNNs could be easier to determine the model values from both patterns and location information together than relying only on possible indistinguishable patterns. Finally, we let CNNs learn the mapping from data and location to model value

$$m = \mathcal{F}(d, t) \quad (4)$$

where $t$ denotes the tier positions in vertical directions of apparent resistivity data. In the following section, we will detail the architecture of CNNs and how we introduce $t$.

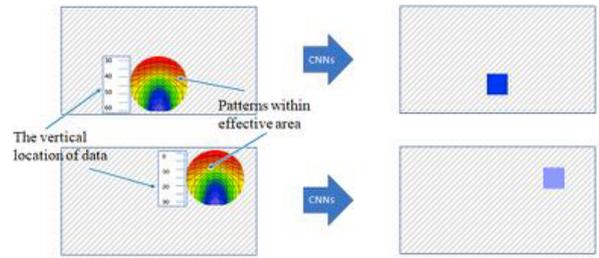

Fig. 5. Illustration of how we reduce the potential ambiguity of CNNs during training. When patterns themselves are not distinguishable, we supply them with the corresponding vertical location information to help build the mapping.

### B. Networks

We design our networks based on prevalently used U-Net architecture [40] as shown in Fig. 6. U-Net is well known for its shortcut operation which concatenates feature maps from the shallow layer (low-level feature maps) to feature maps from the deep layer (high-level feature maps). Normally, high-level features contain knowledge more related to the final result value, whereas low-level features have knowledge related to some general concepts such as position and shape. In this way, the shortcut would make the last several layers give outputs based on high-level and low-level knowledge together, so as to help get final results with both accurate value and anomalous morphology. Moreover, the shortcut will help backpropagate gradients and accelerate parameter optimization in shallow layers. We also add several residual blocks [24] at the end of U-Net to enhance the capacity. Finally, there are 22 layers with convolution operation ($3 \times 3$ kernel, 1 stride), 4 layers with transposed convolution (fractionally strided convolution) operation ($2 \times 2$ kernel, 0.5 stride), and also 4 layers of max-pooling operation ($2 \times 2$ kernel, 2 stride), which results in large enough receptive field ($238 \times 238$) and nonlinearity for our data and task. (For a detailed calculation of the receptive field the reader can refer the Supplementary Material.)

To reduce the potential ambiguity when applying CNNs in our task as discussed in Section III-A, we introduce tier feature map and concatenate it to the input data to supplement tier position information. In typical surface detection scenarios, the apparent resistivity data with different tier positions in the vertical direction can be obtained by changing the distance between injecting electrodes, which is the basic method of electrical sounding. Under different sounding conditions, the electrode device moves horizontally along the survey line to form apparent resistivity profile, thereby the data matrix of apparent resistivity could be acquired. In the data matrix of apparent resistivity, the data with different tier positions are strongly correlated with the anomalous structures of corresponding depths. That is to say, introducing tier position information into apparent resistivity data can be regarded as supplementing depth information for CNNs. Our tier feature map is with tier structure that element in each tier has a value equal to the tier number and has the same spatial dimension as $d$. We denote tier feature map as $t$ and $t$ have the same dimension as the data $d$. Each element of $t$ is defined as $t_{i,j} = i$, where $i, j$ indicate vertical and horizontal locations, respectively, as shown in Fig. 7(a). Thus, elements of $t$ in the same horizontal locations have the same value. For this reason, we name $t$ as tie feature map. When the data $d$ is of height $h$, $t$ will have $h$ tiers. Concatenating $t$ to the input data is equivalent to treat $t$ as another channel of the input data, as illustrated in Fig. 7(b).

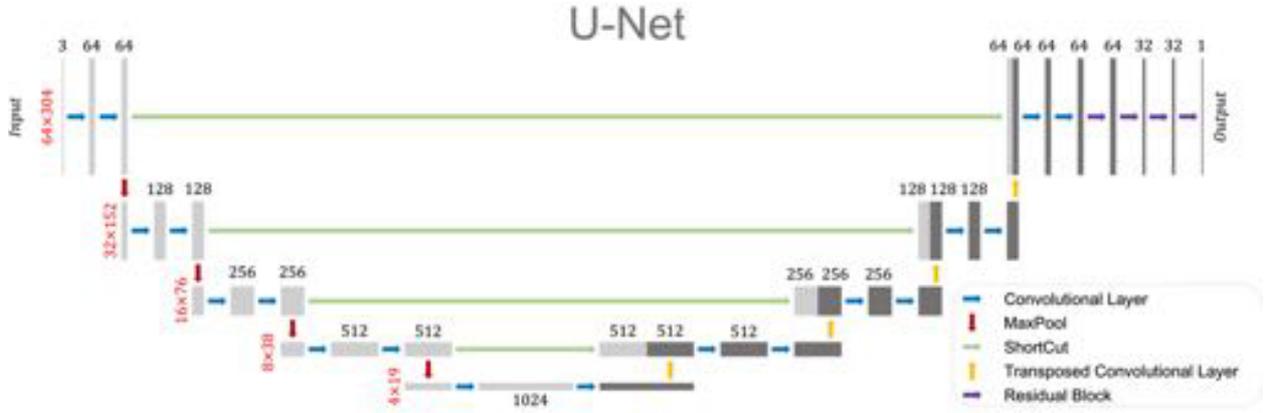

Fig. 6. U-Net architecture. The black number above feature maps indicates the number of channels of feature maps while the red number beside feature maps denotes their spatial dimension. Blue arrow represents convolutional layer while purple arrow introduces extra residual connection. Red arrow means max-pooling operation which downsamples feature maps while yellow arrow means transposed convolution which upsamples feature maps. Green arrow is the shortcut which concatenates feature maps from shallow layers to deep layers.

### C. Loss Function

*1) Basic Metric and Weighting Function:* For the value regression problem, in loss function, we apply prevalently used mean square error (MSE) metric for data value term $v$ to penalize the misfit of inversion value $\hat{m}_{i,j}$ at position $(i, j)$ with respect to the groundtruth $m_{i,j}$, and we assume $d$ and $m$ have spatial dimension of $[H, W]$, thus the data value term is defined as

$$v(\hat{m}, m) = \sum_{i,j}(\hat{m}_{i,j} - m_{i,j})^2. \qquad (5)$$

For classical ERS inversion, it is usually more difficult to obtain accurate inversion results for deep anomalies.

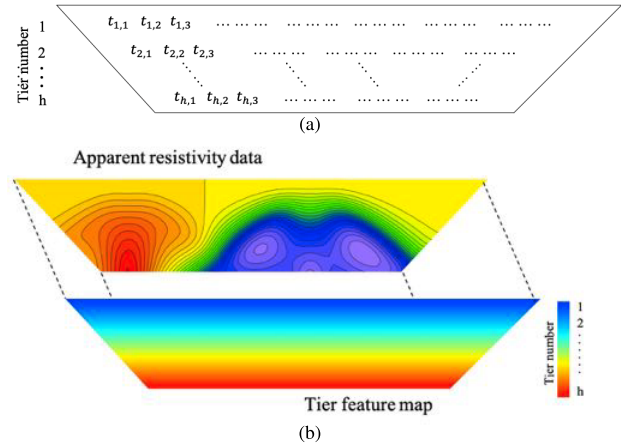

Fig. 7. (a) Tier feature map is with tier structure that element in each tier has the value equal to tier number. (b) Our actual input data after concatenating tier feature map.

Li and Oldenburg [42], [43] proposed a weighting function to counteract the natural decay of the static field to overcome the tendency of putting structure at the surface. Its effectiveness has been demonstrated [44], [45]. In this case, we also take the idea of depth weighting function in the loss function to let the network invest with more capability in the deep area. In this way, the inversion accuracy and resolution of deep anomalous bodies will be improved. The depth weighting function $dw$ is defined as

$$dw(\hat{m}_{i,j}) = (i + \lambda)^{\beta/2} \qquad (6)$$

where $\lambda$ is the constant parameter related to the grid size and the location of the current electrodes, and the parameter $\beta$ is a constant for controlling depth weight distribution. The configuration of these two parameters will be provided in Section IV-B.

Finally, with depth weighting function $dw$, we design our data value term $v$ as

$$v(\hat{m}, m) = \sum_{i,j} dw(\hat{m}_{i,j}) \cdot (\hat{m}_{i,j} - m_{i,j})^2. \qquad (7)$$

*2) Smooth Constraints:* Inversion tasks are often mathematically ill-posed that the solutions are usually nonunique and unstable. One way to solve this problem is by adopting the well-tested smoothness constrained least-squares approach [46]. Restricted by the smooth constraints, sudden changes between adjacent grids in the resistivity model will be reduced. We carry out the smooth constraints by introducing smooth term

$$s(\hat{m}) = \sum_{i,j} |\hat{m}_{i+1,j} - \hat{m}_{i,j}| + |\hat{m}_{i,j+1} - \hat{m}_{i,j}|. \tag{8}$$

The smooth term plays the role of regularization and is also known as total variation loss.

*3) Final Formulation:* Consequently, our final loss function $L$ is defined as

$$L = \frac{1}{Z}(v(\hat{m}, m) + \alpha \cdot s(\hat{m})) \tag{9}$$

where $\alpha$ is the smoothness factor (will be decided and discussed in Sections IV-B and V-C), and $Z = HW$ is the normalization term. All the operations and losses are derivable and result in our end-to-end networks which we call ERSInvNet.

It is worth noting that though the backbone of networks is the prevalent U-Net initially used for computer vision tasks, with our proposed tier feature map and depth weighting function, the networks have been converted to our ERSInvNet which is specific for apparent resistivity data with vertically varying characteristic.

## IV. EXPERIMENTS

### A. Data Set Preparation

For deep learning-based geophysical inversion, data set generally should reach a sufficient amount and guarantee the diversity. As such, in our work, we collect a data set with 36 214 sample pairs, which is called ERSInv data set. Each sample pair consists of one resistivity model (with spatial dimension [64 × 304]) and one corresponding apparent resistivity data (with spatial dimension [64 × 304]). Resistivity model is designed by referring to real 2-D ERS scenarios. We generate synthetic data by predefining a few anomalous bodies with different resistivity value, and then embedding them to the different positions of homogeneous medium (500 Ω · m). The resistivity anomalous bodies consist of 5 subsets as follows. *Type I*: Single rectangular body (5236 sample pairs), *Type II*: Two rectangular bodies (7560 sample pairs), *Type III*: Three rectangular bodies (7920 sample pairs), *Type IV*: Single declining bodies (6426 sample pairs), and *Type V*: Two declining bodies (9072 sample pairs). For each type, the resistivity anomalous bodies may have different resistivity values, in our data set low resistivity anomaly is the one with the body value from [10 Ω · m, 20 Ω · m, 50 Ω · m] and high resistivity anomaly is the one with body value from [1000 Ω·m, 1500 Ω·m, 2000 Ω·m]. Accordingly, we have a total of five different types of resistivity models and the corresponding apparent resistivity data are generated by forward modeling. Schematic and parameters of anomalous bodies are shown in Table I.

The selection of electrode configuration for the ERS is crucial in acquiring the response of the observed target because different electrode configurations have different horizontal and vertical resolution [47]. The apparent resistivity data of Wenner and Wenner–Schlumberger arrays are adopted in this work since these two configurations will have good vertical resolution and appropriate horizontal resolution [48], [49] when they used together. Our observation data are generated through forward modeling on resistivity models. Specifically, the finite element approach with uniform square grid [50] is applied for forward modeling.

The simulated electrical fields are usually generated by finite-element methods based on the anomalous potential method. The values of both input and output during training are normalized to the range of [0, 1] since standardizing either input or target variables tend to make the training process better behaved [19]. We use two kinds of apparent resistivity data (Wenner and Wenner–Schlumberger configurations) and the tier feature map proposed in Section III-B to form the input data of three channels. The data set is randomly divided into training set, validation set, and test set in a ratio of 10 : 1 : 1 (training set: 30 180 pairs; validation set: 3017 pairs; test set: 3017 pairs).

### B. Implementation

The neural networks in this work are built using PyTorch. SGD optimizer with batchsize 5, learning rate 0.1, momentum 0.9, and weight decay 1$e$-4 is used to optimize networks. During training, we carry out 500 epochs of optimization in total, and also perform one time of validation after each training epoch to verify the training effect. Following Li and Oldenburg [42], [43], the parameters $\beta$ and $\lambda$ of depth weighting function are set to 1 and 8, respectively, and the smoothness factor $\alpha$ is experimentally set to 0.2. A study of influence of $\alpha$ is carried out in Section V-C. In this work, the hyperparameters such as $\beta$, $\lambda$, and $\alpha$ are chosen according to the evaluation on the validation set. All computations are carried out with the machine of single NVIDIA TITAN Xp. It is worthy to note that in this environment, our ERSInvNet could reach very fast inference speed during testing with 0.013 $s$/data.

In order to verify the proposed ERSInvNet, six experiments are arranged as follows: *Experiment 1*, ERSInvNet performance analysis; *Experiment 2*, the ablation study of tier feature map; *Experiment 3*, the ablation study of depth weighting function and smooth regularization; *Experiment 4*, results on the noisy data; *Experiment 5*, comparison with iterated linear method; *Experiment 6*, one field test. All the ablation studies are carried out in test and validation sets.

In addition to qualitative evaluation through visual judgment, weighted mean square error (*WMSE*) and weighted correlation coefficient (*WR*) are also used to measure the statistic performance quantitatively, which is given as follows:

$$\text{WMSE} = \frac{1}{N} \sum_{n=1}^{N} [w^n(\hat{m}^n - m^n)]^T [w^n(\hat{m}^n - m^n)]$$

$$\text{WR} = \frac{1}{N} \sum_{n=1}^{N} \frac{[w^n(\hat{m}^n - \bar{\hat{m}}^n)]^T [w^n(m^n - \bar{m}^n)]}{|w^n(\hat{m}^n - \bar{\hat{m}}^n)|_2 |w^n(m^n - \bar{m}^n)|_2} \tag{10}$$

TABLE I
SCHEMATIC AND PARAMETERS OF ANOMALOUS BODIES

| 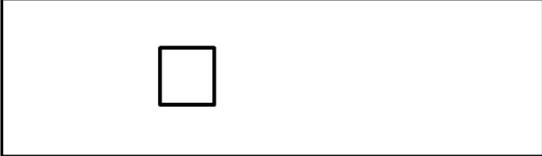 | •Type I: Single rectangular body<br>•Model quantity: 5236 pairs<br>•Anomalous bodies size: 4×4; 6×6; 8×8; 10×10; 12×12; 14×14; 16×16; 18×18; 20×20; 8×30; 20×10<br>•Anomalous bodies types: 1 low resistivity body; 1 high resistivity body |
|---|---|
| 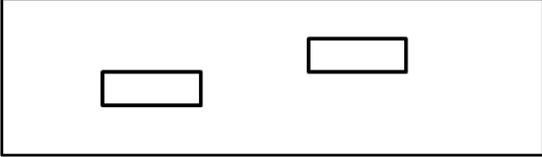 | •TypeII: Two rectangular body<br>•Model quantity: 7560 pairs<br>•Anomalous bodies size: 8×8; 10×10; 8×30; 20×10<br>•Anomalous bodies types: 2 low resistivity bodies; 2 high resistivity bodies; 1 low resistivity body & 1 high resistivity body |
| 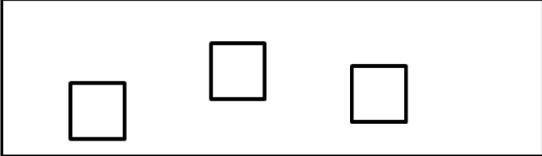 | •TypeIII: Three rectangular body<br>•Model quantity: 7920 pairs<br>•Anomalous bodies size: 8×8; 10×10; 8×30; 20×10<br>•Anomalous bodies types: 3 low resistivity bodies; 3 high resistivity bodies; 1 low resistivity body & 2 high resistivity body; 2 low resistivity body &1 high resistivity body |
| 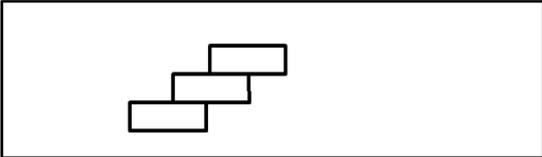 | •Type IV: Single declining body<br>•Model quantity: 6426 pairs<br>•Anomalous bodies size: 8×4; 10×5; 12×6  layer: 3; 4; 5<br>•Anomalous bodies types: 1 low resistivity body; 1 high resistivity body |
| 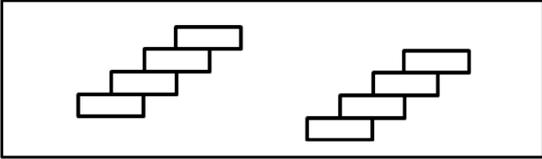 | •Type V: Two declining body<br>•Model quantity: 9072 pairs<br>•Anomalous bodies size: 8×4; 12×6  layer: 4; 5<br>•Anomalous bodies types: 2 low resistivity bodies; 2 high resistivity bodies; 1 low resistivity body & 1 high resistivity body |

where $m^n$ and $\hat{m}^n$ are vectorized actual and predicted model values, respectively, whereas $w^n$ is the vectorized weight, and we use $\bar{\cdot}$ to denote the average values of $\cdot$. $w^n$ is designed to make the region far from anomalies in the resistivity model that has large weight because false anomalies far from true anomalies are not preferred, whereas false anomalies closed to true anomalies are usually acceptable. $N$ is the number of samples. $|\cdot|_2$ indicates $L_2$ norm. *WMSE* measures the value fitting between prediction and groundtruth with the value the lower the better, while *WR* measures the statistical relationship between prediction and groundtruth with the value the larger the better. We also show *WMSE* and *WR* for all the examples we used in the experiments.

## V. RESULTS AND DISCUSSION

### A. Results of Experiment 1

In Experiment 1, some examples demonstrate the inversion performance of the proposed method. The misfit degree of locations and shapes of anomalous bodies as well as resistivity values are the major factors considered during evaluation. In Fig. 8, we randomly select five inversion results that correspond to five model types, respectively, in the test set. The images arranged from left to right are the corresponding groundtruth, apparent resistivity data, ERSInvNet results, and the vertical resistivity profiles, respectively.

From the first two columns, we can note that spatial correspondence extensively exists between the apparent resistivity data and the resistivity models. From the overall observation of Fig. 8, ERSInvNet could accurately predict the model value and also get good localization of anomalous bodies, which demonstrate its promising inversion ability. In order to visualize the positions, shapes, and resistivity values of anomalous bodies in inversion results more intuitively, the resistivity change along the anomalous body profile (shown by the black line in Fig. 8) is given in the form of curves on the fourth column. We can see that the resistivity curves of inversion results and models are almost aligned (with the relative error within 0.4%, the relative error is defined as |predict-real|/real) and change synchronously in most places except for the regions near the boundary of the anomalous bodies. This is because the smooth constraints restrict the mutation of resistivity value near the abnormal body boundary. The effect of smooth constraints will be discussed in Section V-C.

In the third example, three anomalous bodies with different depths can be clearly and accurately reflected in the inversion results. Among them, the high resistivity bodies at a depth of 15 and 25 m are closed to the model value, whereas the resistivity of the deepest one (with value 1400 $\Omega \cdot$ m) is lower than the corresponding high resistivity body in the model (with value 2000 $\Omega \cdot$ m). The reason is that when powered on the surface, the field responses caused by deep anomaly have not shown obvious patterns in the apparent resistivity data. The lack of obvious pattern makes ERSInvNet hard to give accurate predictions.

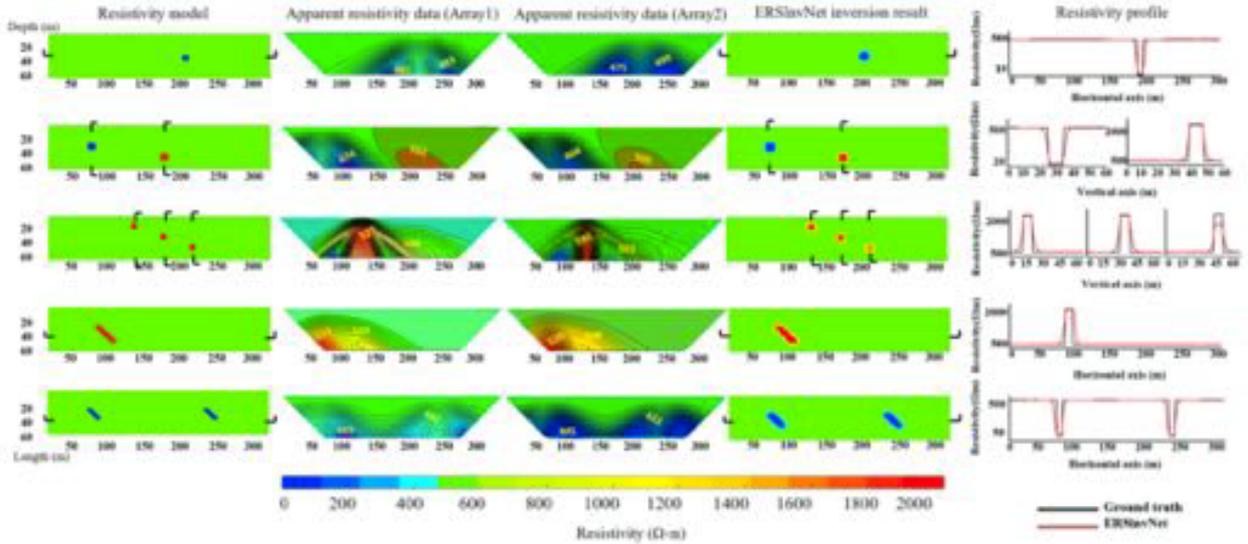

Fig. 8. Groundtruth (first column), inversion results of our ERSInvNet (fourth column) and corresponding apparent resistivity data (second and third columns) on the test set. Vertical and horizontal profiles of resistivity models indicated by the truncation line are shown in the rightmost column for comparing the inverted resistivity values. Rows from top to bottom exhibit examples with the anomalous bodies from *type I–V*, respectively. The yellow numbers in the second and third columns denote the values of apparent resistivity.

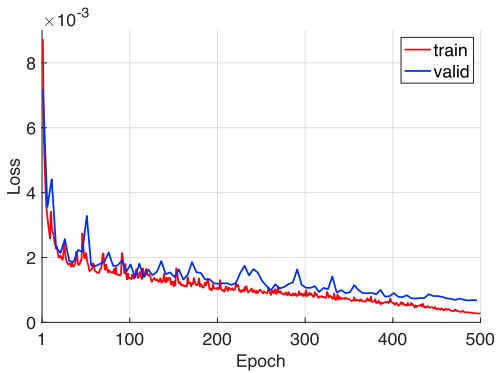

Fig. 9. Loss curves of ERSInvNet on training and validation sets.

In Fig. 9, we show the loss curves of our ERSInvNet on the training and validation set. Both loss curves decrease gradually with the increase of epochs, which indicates the nonexistence of overfitting during training. When the epochs reached 500 times, the loss was reduced from 0.008 to 0.001 and the trend of decline seems will continue.

### B. Results of Experiment 2

In Experiment 2, to verify the role of tier feature map, we compare the ERSInvNet results with/without the tier feature map when having both depth weighting function and smooth constraints in the loss function. Examples with five different types of anomalous bodies are randomly selected. Inversion results with/without tier feature map are shown in Fig. 10. From Example 1, Example 2, and low resistivity body on the right side of Example 5, we can see that the supplement of tier information helps improve the morphological accuracy when inverse resistivity anomalies exist. In addition, the delineation of the boundaries of anomalous bodies is also improved. In Example 2 and Example 3, for multiple anomalous bodies, the tier feature map helps suppress the obvious false anomalies near true ones. Specifically, in Example 2, the false anomalies around the low resistivity anomalous body are removed after introducing the tier feature map, meanwhile the shape of the anomalous body is more accurate. Similarly, three obvious high resistivity false anomalies in Example 3 are completely removed.

From Fig. 10, the ambiguity problem discussed in Section III is verified that there are many false anomalies in the output without the tier feature map. As for what makes the ambiguous output looks like is dependent on what samples the networks learn during training. Since the networks may become confused for some data without a tier feature map, thus reducing the overall loss on the whole training set at best, the output will look like a combination of many possible results. These possible results are from the situations the networks had been encountered.

In summary, the tier feature map can suppress false anomalies. Such rules generally exist in other results. To check the overall performance on the validation and test sets, we quantitatively compare results by *WMSE* and *WR* metrics in Table II. It is easy to understand that quantitative evaluation also supports the positive effect of the tier feature map. Certainly, besides the effects of the tier feature map, the inversion performance also depends on the contribution of our smooth constraints and depth weighting function, which will be discussed in the next section.

### C. Results of Experiment 3

In Experiment 3, we compare the results with/without smooth constraints and depth weighting function when having tier feature map in the input data. Thus, we have four configurations in total that with smooth constraint and depth weighting function together (*SD*), with only smooth constraint (*OS*), with only depth weighting function (*OD*) and with nothing (*NA*). In Fig. 11, resistivity models and inversion results of *NA*, *OS*, *OD*, and *SD* are given from left to right.

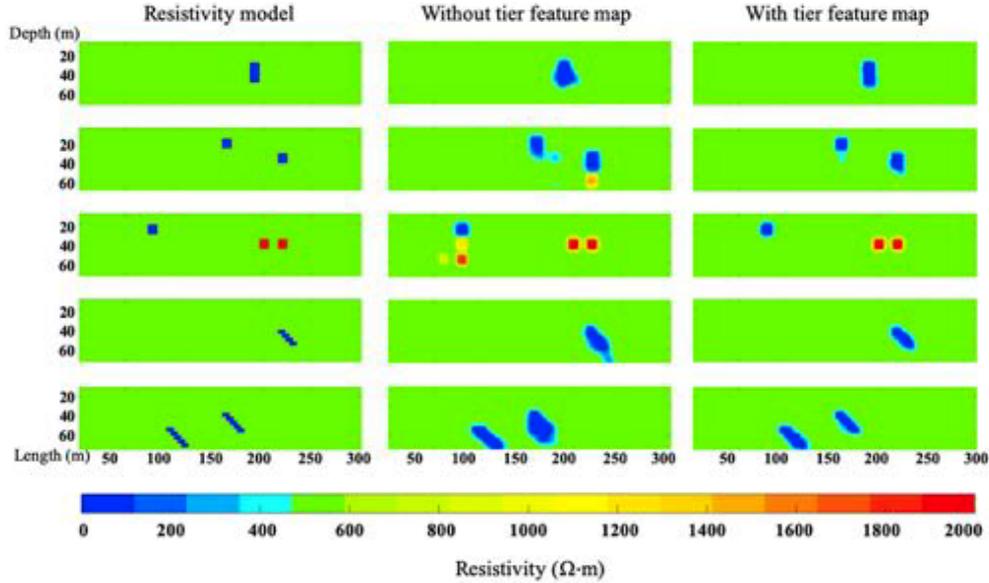

Fig. 10. ERSInvNet inversion results with/without tier feature map on the test set. Rows from top to bottom exhibit examples with the anomalous bodies from *type I–V*, respectively. The comparison in this figure indicates that the tier feature map has positive effects on the inversion accuracy.

TABLE II
QUANTITATIVE COMPARISON OF INVERSION RESULTS WITH/WITHOUT TIER FEATURE MAP ON THE TEST AND VALIDATION SETS. THE ↑ INDICATES THE LARGER VALUE ACHIEVED, THE BETTER PERFORMANCE IS, WHILE ↓ INDICATES THE SMALLER, THE BETTER

|  | Test | | Valid | |
|---|---|---|---|---|
|  | WMSE↓ | WR↑ | WMSE↓ | WR↑ |
| With tier feature map | 0.000335 | 0.540852 | 0.000341 | 0.538876 |
| Without tier feature map | 0.000731 | 0.387227 | 0.000721 | 0.385944 |

By comparing the results of *NA* and *OS*, we see the results with smooth constraints have fewer false anomalies but poor boundary accuracy. (See the second and third columns.) By comparing *NA* and *OD*, we found results with depth weighting function have more accurate anomaly morphology as well as anomaly value, especially in the deep area. That is to say, the main contribution of smooth constraints is to suppress false anomalies, while depth weighting function will benefit inversion accuracy. The overall comparison indicates that ERSInvNet with all the tier feature map, smooth constraints, and depth weighting function (*SD*) has the best performance.

*WMSE* and *WR* scores of four different ERSInvNet configurations on the validation and test sets are listed in Table III. On the whole, *SD* has the highest *WR* as well as the lowest *WMSE* value, which is consistent with the results of the visual comparison. However, *OD* and *OS* are unexpected and even worse than *NA* after introducing smooth constraints and depth weighting function. In particular, when smooth constraints are applied in *OS*, the *WMSE* increases by 74.7%. It is because smooth constraints would sacrifice much accuracy on delineating the boundaries of anomalies. For the performance decrease of only applying depth weighting function in *OS*, we guess it may be caused by the overproduced false anomalies in the deep area of inversion results. To be specific, with depth weighting function, failed identification of some anomalies will cause a significant increase of loss, and thus to avoid this, networks may tend to produce more false anomalies around the real ones. After introducing smooth constraints and depth weighting function together (*SD*), we obtained the best performance that indicates that smooth constraints and depth weighting function can mutually benefit and restrain the negative effects.

In Fig. 12, we further report on the loss curves of ERSInvNet with different smoothness factor $\alpha$ on the validation set, which demonstrates certain robustness of our method to the hyperparameter choosing and setting $\alpha$ to 0.2 yield the better performance.

### D. Results of Experiment 4

The measured data set always contains noise in practice. Thus, in Experiment 4, we add random noise in the synthetic input to study how the noise impact on the output. We add the white Gaussian noise to the input with the intensity 1 and 3 dBw. The example figure of data after adding noise is given in the Supplementary Material. In Fig. 13, inversion results from synthetic observed data without noise, with 1-dBw noise, and with 3-dBw noise are given from left to right. The five sets of examples are selected as the same as the ones shown in Fig. 8.

As can be seen from Fig. 13, the inversion results received a different degree of influence for different degree of noise. For the example in the first row, the shape and location are basically accurate under the influence of 1-dBw noise, but the size of the anomalous body becomes much smaller. When the noise is increased to 3 dBw, there appears an offset for the position of the anomalous body. The inversion results of the low resistivity body in the example of the second row are not much affected by noise, which may be due to its shallower burial depth and larger volume. This assumption is also confirmed by the example in the fifth row, where two larger inclined low resistivity bodies do not show the large difference at different noise degree. However, the inversion effect of high resistivity bodies is significantly affected by noise. As examples in the second row, the high resistivity body is missed under the influence of noise. This phenomenon also appears in the third row where the high resistivity body in the deepest position disappears from the inversion results of noisy data, whereas the other two high resistivity bodies are still accurately inverted. For inversion results in the fourth row, the upper boundaries of the high resistivity region are all accurately performed, but the deeper parts are missed under the influence of noise.

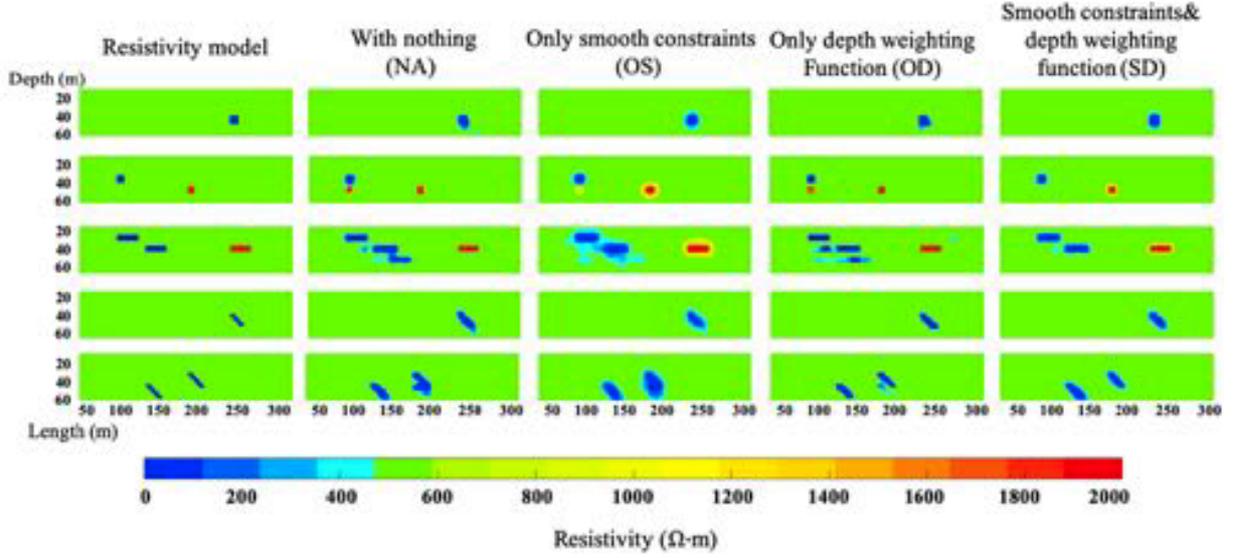

Fig. 11. Inversion results on different configurations on the test set. Rows from top to bottom exhibit examples with the anomalous bodies from *type I–V*, respectively. The comparison indicates that the smooth constraints could suppress false anomalies and depth weighting function benefits the inversion accuracy.

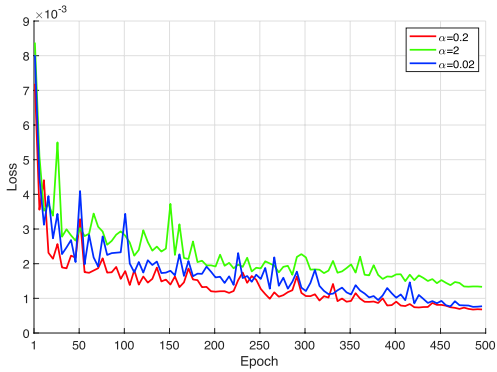

Fig. 12. Loss curves of ERSInvNet with different smoothness factor $\alpha$ on validation set.

TABLE III
QUANTITATIVE COMPARISON OF INVERSION RESULTS WITH/WITHOUT SMOOTH CONSTRAINTS AND DEPTH WEIGHTING FUNCTION ON THE TEST AND VALIDATION SETS. THE ↑ INDICATES THE LARGER VALUE ACHIEVED, THE BETTER PERFORMANCE IS, WHILE ↓ INDICATES THE SMALLER, THE BETTER

|    | Test WMSE↓ | Test $WR$ ↑ | Valid WMSE↓ | Valid $WR$ ↑ |
|----|------------|-------------|-------------|--------------|
| SD | 0.000335   | 0.540852    | 0.000341    | 0.538876     |
| OD | 0.000608   | 0.387756    | 0.000592    | 0.387921     |
| OS | 0.000959   | 0.335904    | 0.000971    | 0.335029     |
| NA | 0.000549   | 0.425744    | 0.000555    | 0.421256     |

Generally, the deeper the depth and the smaller the volume of the anomalous body, the more influence the inversion results received from noise. At the same time, the high resistivity body is much affected than the low resistivity body. That is to say, anomalous bodies which cause smaller anomalous amplitude in observed data are more susceptible to noise. The quantitative comparison of results under different degree of noise is provided in Table IV, where each metric value is the average among the five examples in Fig. 13. In order to improve the antinoise ability of deep learning inversion, we will introduce the idea of transfer learning [51], [52] in future research, which will adapt the trained networks to noise data by fine-tuning the parameters.

### E. Results of Experiment 5

In Experiment 5, we benchmark ERSInvNet against the well-known iterated linear inversion using *RES2DINV* software which is widely applied in ERS inversion. For a fair comparison, we use synthetic model with position, size, and resistivity value of anomalies that are unprecedented during the training of our ERSInvNet. And the same configurations are used to generate corresponding resistivity data for both methods. Fig. 14(a) and (b) shows two examples in which (i) synthetic resistivity model, (ii) inversion result of the iterated linear method *RES2DINV*, and (iii) inversion result of the proposed ERSInvNet are shown. The results of both examples can depict the existence of anomalous bodies. Meanwhile, it can be seen that the ERSInvNet predicts the location, shape of the conductive block, and the resistivity value more precisely. Quantitative comparison given in Table V also verifies the superiority of the proposed ERSInvNet. Compared with traditional methods, DNN-based methods could utilize data prior learned from the training set as well as human-introduced priors such as smoothness, meanwhile they have more powerful nonlinear approximation abilities. With all these advantages, DNNs reach these promising results in our task.

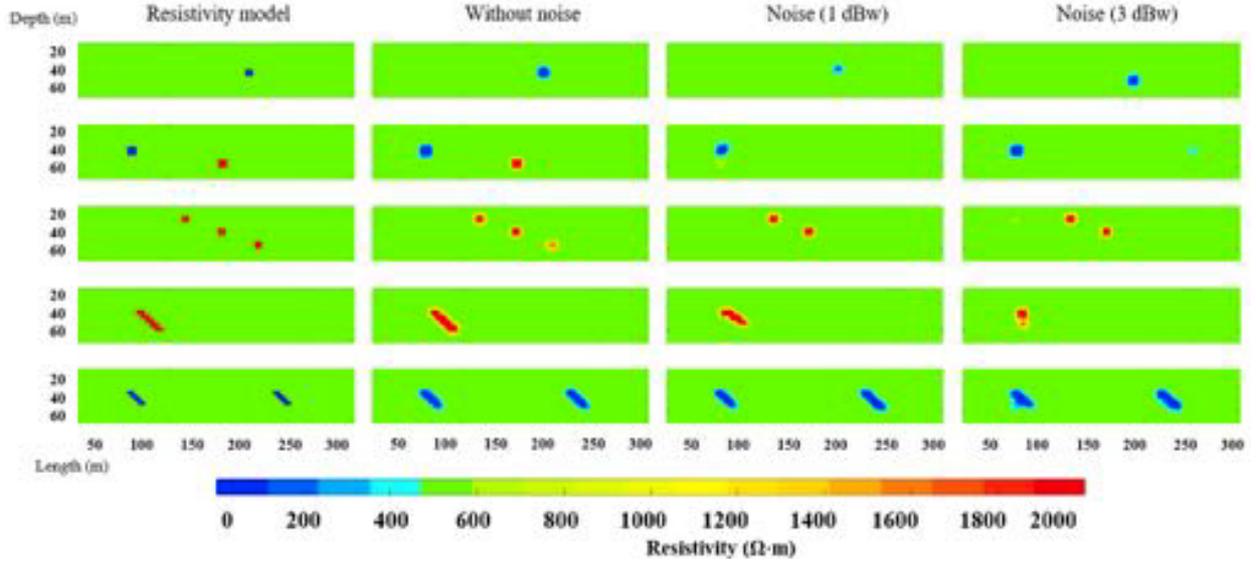

Fig. 13. Inversion results from pure and noisy synthetic observed data. (From left to right) Groundtruth (the first column), inversion results without noise (the second column), inversion results with 1 (dBw) noise (the third column), and inversion results with 3 (dBw) noise (the forth column) from the test set. (From top to bottom) each row exhibits one example from the anomalous type $I$–$V$, respectively.

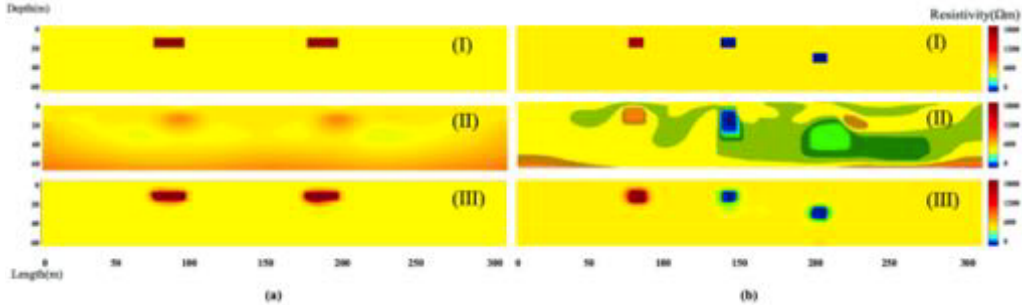

Fig. 14. Synthetic models and inversion results. (a) and (b) Two examples in which (i) synthetic models, (ii) inversion results of the iterated linear method, and (iii) inversion results of the proposed ERSInvNet.

TABLE IV
QUANTITATIVE COMPARISON OF RESULTS UNDER DIFFERENT DEGREE OF NOISE. THE ↑ INDICATES THE LARGER VALUE ACHIEVED, THE BETTER PERFORMANCE IS, WHILE ↓ INDICATES THE SMALLER, THE BETTER

|  | WMSE↓ | WR↑ |
|---|---|---|
| Without noise | 0.000221 | 0.985997 |
| Noise (1 dBw) | 0.001568 | 0.720364 |
| Noise (3 dBw) | 0.002482 | 0.513130 |

### F. Results of Field Test

In this experiment, we use ERSInvNet trained on the proposed ERSInv Data set with noise and with the proposed tier feature map and loss function defined in Sections III-B and III-C. The survey was conducted at the waterproof curtain grouting area of Dehou reservoir project, Yunnan Province, China. According to the preliminary geological investigations, the construction area is located in the bedrock mainly dominated by carbonate rock. The karst system developed in the rock mass (karst caves, corrosion fractures, etclet@tokeneonedot.) may lead to water leakage in the reservoir area. Thus, the field data were from the survey, which was carried out to determine the subsurface water-bearing karst structures for the water leakage investigation in the reservoir area.

TABLE V
QUANTITATIVE COMPARISON OF INVERSION RESULTS ON TWO EXAMPLES IN FIG. 14. THE ↑ INDICATES THE LARGER VALUE ACHIEVED, THE BETTER PERFORMANCE IS, WHILE ↓ INDICATES THE SMALLER, THE BETTER

| | (a) | | (b) | |
|---|---|---|---|---|
| | WMSE↓ | $WR$ ↑ | WMSE↓ | $WR$ ↑ |
| ERSInvNet | 0.000928 | 0.964870 | 0.000293 | 0.980529 |
| RES2DINV | 0.559029 | 0.133533 | 0.546130 | 0.230070 |

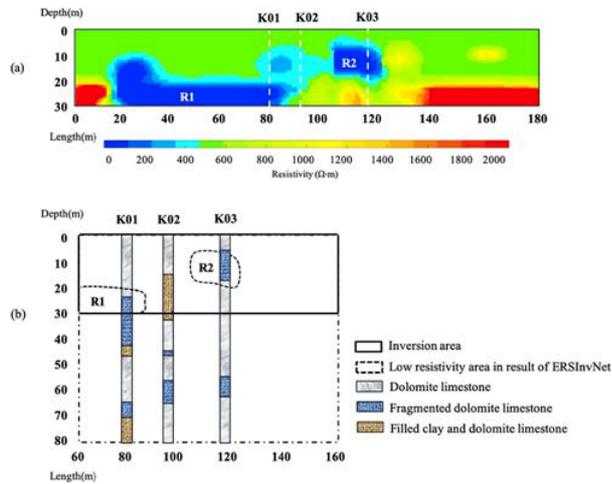

Fig. 15. Field test results. (a) Inversion result of the proposed ERSInvNet. (b) Geological drillings results in study area.

One 2-D electrical resistivity tomography (ERT) (2-D-ERT) profile was arranged along the study area. AGI SuperSting system [53] was used for data collection. The length of the survey line is 180m, and the electrode spacing is 3 m. Fig. 15 displays the inversion results of ERSInvNet. Three vertical boreholes $K01$, $K02$, and $K03$ had been conducted in the study area and were located at 82m, 94m, and 118m along the survey line [as shown in the white dotted line of Fig. 15(a)].

The inversion result shows that two low resistivity areas exist and the minimum resistivity is less than 10 $\Omega \cdot$ m. $R1$ is located at 21-89 m (length) and 18-30 m (depth), whereas $R2$ is located at 111-124 m (length) and 6-19 m (depth). $R1$ area could be a suspected aquifer in dolomite limestone strata, and $R2$ area could be inferred as a water-bearing karst cave. The geological drilling results are shown in Fig. 15(b). Fragmented dolomite limestone area was exposed at a depth of 23–42 m in $K01$ and at a depth of 5–17 m in $K03$. The low resistivity areas (lower than 10 $\Omega \cdot$ m) were basic coincidence with the drilling results. According to the interpretation, we inferred that the study area had a high possibility of water leakage and grouting work should be carried out appropriately.

## VI. CONCLUSION

In this article, we propose a CNN-based network called ERSInvNet for inverse problems on resistivity data. Though some attempts of CNN-based tomography have been made, ERS inversion is different from the previous studies because of the vertically varying characteristic inherent in apparent resistivity data. This characteristic will lead to ambiguity when using CNNs directly. To address this issue, we supplement a tier feature map to the input data. Besides, to further reduce the false anomalies and improve the prediction accuracy for the deep region, smooth constraints and depth weighting function are introduced into loss function during training.

To train, validate, and test the proposed ERSInvNet, we collect a synthetic data set that contains 36 214 pairs of apparent resistivity data and resistivity model. Comparative experiments show that including the tier feature map helps obtain more accurate inversion results and suppress false anomalies. The individual use of smooth constraints and depth weighting function can reduce false anomalies or improve prediction accuracy for the deep region. However, it will sacrifice performance in other aspects. Then, simultaneously using both of them could mutually benefit and achieve the best results, which are demonstrated through comprehensive qualitative analysis and quantitative comparison. Moreover, comparing with traditional methods, ERSInvNet could reach a very fast inference speed during testing and more accuracy inversion results. Finally, a series of experiments especially the one on the field data demonstrated the promising performance and significant potential of ERSInvNet for the resistivity inversion task.

## REFERENCES

[1] M. Loke, J. Chambers, D. Rucker, O. Kuras, and P. Wilkinson, "Recent developments in the direct-current geoelectrical imaging method," *J. Appl. Geophys.*, vol. 95, pp. 135–156, Aug. 2013.

[2] D. J. Labrecque, A. L. Ramirez, W. D. Daily, A. M. Binley, and S. A. Schima, "ERT monitoring of environmental remediation processes," *Meas. Sci. Technol.*, vol. 7, no. 3, pp. 375–383, Mar. 1996.

[3] K. Hayley, L. Bentley, and M. Gharibi, *Time-Lapse Electrical Resistivity Monitoring of Salt-Affected Soil and Groundwater*, document Water Resources Research 45, W07425, 2009.

[4] J. Reynolds, *An Introduction to Applied and Environmental Geophysics*, 2nd ed. Hoboken, NJ, USA: Wiley, 2011.

[5] J.-H. Kim, M.-J. Yi, Y. Song, S. Jee Seol, and K.-S. Kim, "Application of geophysical methods to the safety analysis of an earth dam," *J. Environ. Eng. Geophys.*, vol. 12, no. 2, pp. 221–235, Jun. 2007.

[6] P. B. Wilkinson, J. E. Chambers, P. I. Meldrum, D. A. Gunn, R. D. Ogilvy, and O. Kuras, "Predicting the movements of permanently installed electrodes on an active landslide using time-lapse geoelectrical resistivity data only," *Geophys. J. Int.*, vol. 183, no. 2, pp. 543–556, Nov. 2010.

[7] P. Sjödahl, T. Dahlin, and S. Johansson, "Using the resistivity method for leakage detection in a blind test at the Røssvatn embankment dam test facility in Norway," *Bull Eng. Geol. Environ*, vol. 69, no. 4, pp. 643–658, Nov. 2010.

[8] S. Park, "Fluid migration in the vadose zone from 3-D inversion of resistivity monitoring data," *Geophysics*, vol. 63, no. 1, pp. 41–51, Jan. 1998.

[9] L. Slater and A. Binley, "Evaluation of permeable reactive barrier (PRB) integrity using electrical imaging methods," *Geophysics*, vol. 68, no. 3, pp. 911–921, May 2003.

[10] D. Rucker, "A coupled electrical resistivity-infiltration model for wetting front evaluation," *Vadose Zone J.*, vol. 8, no. 2, pp. 383–388, May 2009.

[11] S.-H. Cho, H.-K. Jung, H. Lee, H. Rim, and S. K. Lee, "Real-time underwater object detection based on DC resistivity method," *IEEE Trans. Geosci. Remote Sens.*, vol. 54, no. 11, pp. 6833–6842, Nov. 2016.

[12] J. M. Legault, D. Carriere, and L. Petrie, "Synthetic model testing and distributed acquisition dc resistivity results over an unconformity uranium target from the Athabasca Basin, northern Saskatchewan," *Lead. Edge*, vol. 27, no. 1, pp. 46–51, Jan. 2008.

[13] J. Chambers *et al.*, "Bedrock detection beneath river terrace deposits using three-dimensional electrical resistivity tomography," *Geomorphology*, vols. 177–178, pp. 17–25, Dec. 2012.

[14] X. Liu, F. Liu, J. Chen, Z. Zhao, A. Wang, and Z. Lu, "Resistivity logging through casing response of inclined fractured formation," *IEEE Trans. Geosci. Remote Sens.*, vol. 56, no. 8, pp. 4919–4929, Aug. 2018.

[15] C. Schwarzbach, R.-U. Börner, and K. Spitzer, "Two-dimensional inversion of direct current resistivity data using a parallel, multi-objective genetic algorithm," *Geophys. J. Int.*, vol. 162, no. 3, pp. 685–695, Sep. 2005.

[16] B. Liu, S. Li, L. Nie, J. Wang, X. L, and Q. Zhang, "3D resistivity inversion using an improved genetic algorithm based on control method of mutation direction," *J. Appl. Geophys.*, vol. 87, pp. 1–8, Dec. 2012.

[17] S. P. Sharma, "VFSARES—A very fast simulated annealing FORTRAN program for interpretation of 1-D DC resistivity sounding data from various electrode arrays," *Comput. Geosci.*, vol. 42, pp. 177–188, May 2012.

[18] R. Shaw and S. Srivastava, "Particle swarm optimization: A new tool to invert geophysical data," *Geophysics*, vol. 72, no. 2, pp. F75–F83, Mar. 2007.

[19] G. El-Qady and K. Ushijima, "Inversion of DC resistivity data using neural networks," *Geophys. Prospect*, vol. 49, no. 4, pp. 417–430, Jul. 2001.

[20] A. Neyamadpour, W. A. T. W. Abdullah, S. Taib, and D. Niamadpour, "3D inversion of DC data using artificial neural networks," *Studia Geophysica et Geodaetica*, vol. 54, no. 3, pp. 465–485, Jul. 2010.

[21] S. Maiti, V. C. Erram, G. Gupta, and R. K. Tiwari, "ANN based inversion of DC resistivity data for groundwater exploration in hard rock terrain of western Maharashtra (India)," *J. Hydrol.*, vols. 464–465, pp. 294–308, Sep. 2012.

[22] F. Jiang, L. Dong, and Q. Dai, "Electrical resistivity imaging inversion: An ISFLA trained kernel principal component wavelet neural network approach," *Neural Netw.*, vol. 104, pp. 114–123, Aug. 2018.

[23] M. Al-Abri and N. Hilal, "Artificial neural network simulation of combined humic substance coagulation and membrane filtration," *Chem. Eng. J.*, vol. 141, nos. 1–3, pp. 27–34, Jul. 2008.

[24] K. He, X. Zhang, S. Ren, and J. Sun, "Deep residual learning for image recognition," in *Proc. IEEE Conf. Comput. Vis. Pattern Recognit. (CVPR)*, Jun. 2016, pp. 770–778.

[25] A. L. Maas, A. Y. Hannun, and A. Y. Ng, "Rectifier nonlinearities improve neural network acoustic models," in *Proc. Workshop Deep Learn. Audio, Speech Lang. Process. (ICML)*, 2013, vol. 30, no. 1, p. 3.

[26] S. Ioffe and C. Szegedy, "Batch normalization: Accelerating deep network training by reducing internal covariate shift," in *Proc. 32nd Int. Conf. Mach. Learn.*, 2015, pp. 1–11.

[27] Y. LeCun, Y. Bengio, and G. E. Hinton, "Deep learning," *Nature*, vol. 521, no. 7553, pp. 436–444, 2015.

[28] A. Krizhevsky, I. Sutskever, and G. E. Hinton, "ImageNet classification with deep convolutional neural networks," in *Proc. Adv. Neural Inf. Process. Syst.*, vol. 25. Red Hook, NY, USA: Curran Associates, 2012, pp. 1097–1105.

[29] K. Simonyan and A. Zisserman, "Very deep convolutional networks for large-scale image recognition," 2014, *arXiv:1409.1556*. [Online]. Available: https://arxiv.org/abs/1409.1556

[30] P. Jiang, F. Gu, Y. Wang, C. Tu, and B. Chen, "Difnet: Semantic segmentation by diffusion networks," in *Proc. Adv. Neural Inf. Process. Syst.*, vol. 31. Red Hook, NY, USA: Curran Associates, 2018, pp. 1–10.

[31] C. Dong, C. C. Loy, K. He, and X. Tang, "Image super-resolution using deep convolutional networks," *IEEE Trans. Pattern Anal. Mach. Intell.*, vol. 38, no. 2, pp. 295–307, Feb. 2016.

[32] K. H. Jin, M. T. Mccann, E. Froustey, and M. Unser, "Deep convolutional neural network for inverse problems in imaging," *IEEE Trans. Image Process.*, vol. 26, no. 9, pp. 4509–4522, Sep. 2017.

[33] C. B. Choy, D. Xu, J. Gwak, K. Chen, and S. Savarese, "3D-R2N2: A unified approach for single and multi-view 3D object reconstruction," in *Proc. Eur. Conf. Comput. Vis. (ECCV)*, 2016, pp. 1–17.

[34] M. Araya-Polo, J. Jennings, A. Adler, and T. Dahlke, "Deep-learning tomography," *Lead. Edge*, vol. 37, no. 1, pp. 58–66, Jan. 2018.

[35] Y. Wu, Y. Lin, and Z. Zhou, "InversionNet: Accurate and efficient seismic waveform inversion with convolutional neural networks," in *SEG Technical Program Expanded Abstracts 2018*. Tulsa, OK, USA: Society of Exploration Geophysicists, 2018, pp. 2096–2100.

[36] S. Li *et al.*, "Deep-learning inversion of seismic data," *IEEE Trans. Geosci. Remote Sens.*, to be published.

[37] E. Maggiori, Y. Tarabalka, G. Charpiat, and P. Alliez, "Convolutional neural networks for large-scale remote-sensing image classification," *IEEE Trans. Geosci. Remote Sens.*, vol. 55, no. 2, pp. 645–657, Feb. 2017.

[38] G. Cheng, C. Yang, X. Yao, L. Guo, and J. Han, "When deep learning meets metric learning: Remote sensing image scene classification via learning discriminative CNNs," *IEEE Trans. Geosci. Remote Sens.*, vol. 56, no. 5, pp. 2811–2821, May 2018.

[39] Q. Zhang, Q. Yuan, C. Zeng, X. Li, and Y. Wei, "Missing data reconstruction in remote sensing image with a unified spatial–temporal–spectral deep convolutional neural network," *IEEE Trans. Geosci. Remote Sens.*, vol. 56, no. 8, pp. 4274–4288, Aug. 2018.

[40] O. Ronneberger, P. Fischer, and T. Brox, "U-Net: Convolutional networks for biomedical image segmentation," in *Medical Image Computing and Computer-Assisted Intervention-MICCAI 2015*. Cham, Switzerland: Springer, 2015.

[41] W. Luo, Y. Li, R. Urtasun, and R. Zemel, "Understanding the effective receptive field in deep convolutional neural networks," in *Proc. Adv. Neural Inf. Process. Syst.*, vol. 29, 2016, pp. 4898–4906.

[42] Y. Li and D. Oldenburg, "3-D inversion of magnetic data," *Geophysics*, vol. 61, no. 2, pp. 394–408, 1996.

[43] Y. Li and D. W. Oldenburg, "3-D inversion of gravity data," *Geophysics*, vol. 63, no. 1, pp. 109–119, Jan. 1998.

[44] S. Kang and D. W. Oldenburg, "On recovering distributed IP information from inductive source time domain electromagnetic data," *Geophys. J. Int.*, vol. 207, no. 1, pp. 174–196, Oct. 2016.

[45] P. Qin, D. Huang, Y. Yuan, M. Geng, and J. Liu, "Integrated gravity and gravity gradient 3D inversion using the non-linear conjugate gradient," *J. Appl. Geophys.*, vol. 126, pp. 52–73, Mar. 2016.

[46] A. N. Tikhonov, A. Goncharsky, V. V. Stepanov, and A. G. Yagola, *Numerical Methods for the Solution of Ill-posed Problems*. Norwell, MA, USA: Kluwer, 1995.

[47] Z. Bing and S. Greenhalgh, "Cross-hole resistivity tomography using different electrode configurations," *Geophys. Prospect*, vol. 48, no. 5, pp. 887–912, Sep. 2000.

[48] Y. Sasaki, "Resolution of resistivity tomography inferred from numerical simulation1," *Geophys. Prospect*, vol. 40, no. 4, pp. 453–463, May 1992.

[49] S. Szalai and L. Szarka, "On the classification of surface geoelectric arrays," *Geophys. Prospect*, vol. 56, no. 2, pp. 159–175, Mar. 2008.

[50] P. P. Silvester and R. L. Ferrari, *Finite Elements for Electrical Engineers*, 2nd. ed. Cambridge, U.K.: Cambridge Univ. Press, 1990.

[51] S. J. Pan and Q. Yang, "A survey on transfer learning," *IEEE Trans. Knowl. Data Eng.*, vol. 22, no. 10, pp. 1345–1359, Oct. 2010.

[52] C. Tan, F. Sun, T. Kong, W. Zhang, C. Yang, and C. Liu, "A survey on deep transfer learning," in *Artificial Neural Networks and Machine Learning*. Cham, Switzerland: Springer, 2018.

[53] Y. Maalouf and N. Khoury, "Use of MASW and electrical resistivity in assessing a small earth dam in remote area," in *Proc. 3rd Int. Conf. Adv. Comput. Tools Eng. Appl. (ACTEA)*, Jul. 2016, pp. 78–82.

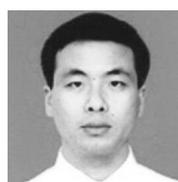

**Bin Liu** received the B.S. and Ph.D. degrees in civil engineering from Shandong University, Jinan, China, 2005 and 2010, respectively.

He joined the Geotechnical and Structural Engineering Research Center, Shandong University, where he is currently a Professor with the School of Qilu Transportation. His research area is engineering geophysical prospecting techniques, especially their applications in tunnels.

Dr. Liu is a member of Society of Exploration Geophysicists (SEG) and International Society for Rock Mechanics and Rock Engineering (ISRM), and serves as a Council Member for Chinese Geophysical Society.

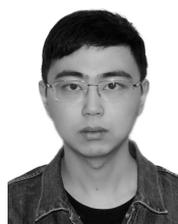

**Qian Guo** is currently pursuing the Ph.D. degree with Shandong University, Jinan, China.

He is mainly engaged in the research work of electrical resistivity detection method.

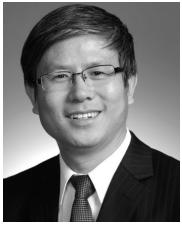

**Shucai Li** received the Ph.D. degree in rock and soil mechanics from the Institute of Rock and Soil Mechanics, Chinese Academy of Sciences, Wuhan, China, in 1996.

He stayed and joined the institute as a Researcher. Since 2001, he has been a Professor with the School of Civil Engineering and the Head of Geotechnical and Structural Engineering Research Center, Shandong University, Jinan, China, where he currently works with the School of Qilu Transportation. He serves as the Chief Editor for Tunneling and Underground Space Technology. His research area is geophysical forward-prospecting of adverse geology during tunneling.

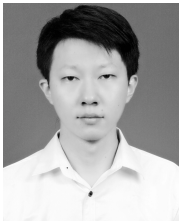

**Benchao Liu** is currently pursuing the master's degree with Shandong University, Jinan, China.

His research interests are mainly deep-learning based on resistivity detection methods.

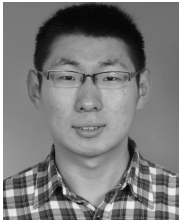

**Yuxiao Ren** received the bachelor's degree in mathematics from Shandong University, Jinan, China, in 2014, and the master's degrees in mathematics from Loughborough University, Loughborough, U.K., in 2015, respectively. He is currently pursuing the Ph.D. degree in civil engineering with Shandong University.

He is currently a Visiting Scholar with the Georgia Institute of Technology, Atlanta, GA, USA, under the supervision of Prof. F. Herrmann. His research interests include seismic modeling and imaging, full-waveform inversion, and deep-learning-based geophysical inversion.

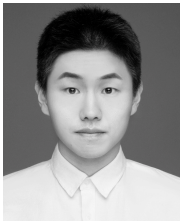

**Yonghao Pang** is currently pursuing the Ph.D. degree with Shandong University, Jinan, China, in 2019.

He is mainly engaged in the research work of geophysical inversion theory.

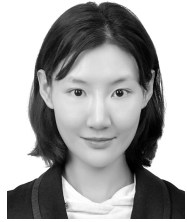

**Xu Guo** received the B.S. degree in financial mathematics from Shandong University, Jinan, China, in 2011, and the Ph.D. degree in applied mathematics from Hong Kong Baptist University, Hong Kong, in 2015.

She then worked as a Post-Doctoral Research Fellow with the University of South Carolina, Columbia, SC, USA, and the Chinese University of Hong Kong. She joined Geotechnical and Structural Engineering Research Center, Shandong University, in 2019, as a Professor. Her research interests focus on mathematical modeling, scientific computing, and data analysis and their applications in tunnels.

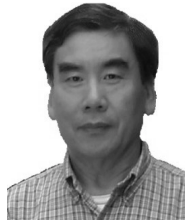

**Lanbo Liu** received the B.S. and M.S. degrees in geophysics from Peking University, Beijing, China, in 1978 and 1981, respectively, and the M.S. degree in civil and environmental engineering and the Ph.D. degree in geophysics from Stanford University, Stanford, CA, USA, in 1992 and 1993, respectively.

He is currently a Professor with the Department of Civil and Environmental Engineering, University of Connecticut, Mansfield, CT, USA. His research interests include numerical modeling and imaging with electromagnetic, acoustic, and seismic waves for exploration, environmental, geotechnical, and biomedical engineering applications.

Dr. Liu served as an Associate Editor for *Geophysics,* and now serves as an Associate Editor for the *Journal of Environmental and Engineering Geophysics*.

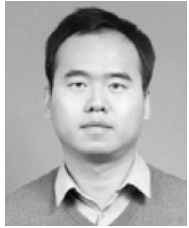

**Peng Jiang** (Member, IEEE) received the B.S. and Ph.D. degrees in computer science and technology from Shandong University, Jinan, China, in 2010 and 2016, respectively.

He is currently a Research Assistant with the School of Qilu Transportation, Shandong University. He has authored or coauthored many works on top-tier venues, including ICCV, NeurIPS(NIPS), the IEEE TRANSACTIONS ON IMAGE PROCESSING, and the IEEE TRANSACTIONS ON GEOSCIENCE AND REMOTE SENSING. Recently, he is focusing on deep learning-based geophysical inversion. His research spans various areas, including computer vision, image processing, machine learning, and deep learning.